\documentclass[final]{cvpr}

\usepackage{times}
\usepackage{epsfig}
\usepackage{graphicx}
\usepackage{amsmath}
\usepackage{amssymb}

\usepackage{booktabs} 
\usepackage{algorithm}
\usepackage[noend]{algpseudocode}
\usepackage{multirow}
\usepackage{wrapfig}
\usepackage[pagebackref=true,breaklinks=true,colorlinks,bookmarks=false]{hyperref}

\def\cvprPaperID{2309} 
\def\confYear{CVPR 2021}

\begin{document}

\title{Sparse Semi-Supervised Action Recognition with Active Learning}

\author{Jingyuan Li\\
Electrical and Computer Engineering\\
University of Washington\\
{\tt\small jingyli6@uw.edu}
\and
Eli Shlizerman\\
Applied Mathematics\\ 
Electrical and Computer Engineering\\
University of Washington\\
{\tt\small shlizee@uw.edu}
}

\maketitle

\begin{abstract}
Current state-of-the-art methods for skeleton-based action recognition are supervised and rely on labels. The reliance is limiting the performance due to the challenges involved in annotation and mislabeled data.
Unsupervised methods have been introduced, however, they organize sequences into clusters and still require labels to associate clusters with actions. 
In this paper, we propose a novel approach for skeleton-based action recognition, called SESAR, that connects these approaches. SESAR leverages the information from both unlabeled data and a handful of sequences actively selected for labeling, combining unsupervised training with sparsely supervised guidance.
SESAR is composed of two main components, where 
the first component learns a latent representation for unlabeled action sequences through an Encoder-Decoder RNN which reconstructs the sequences, and the second component performs \textit{active learning} to select sequences to be labeled based on cluster and classification uncertainty. When the two components are simultaneously trained on skeleton-based action sequences, they correspond to a robust system for action recognition with only a handful of labeled samples. We evaluate our system on common datasets with multiple sequences and actions, such as NW UCLA, NTU RGB+D 60, and UWA3D. Our results outperform standalone skeleton-based supervised, unsupervised with cluster identification, and active-learning methods for action recognition when applied to sparse labeled samples, as low as 1\% of the data.

\end{abstract}

\section{Introduction}

\begin{figure*}
\begin{center}
\includegraphics[width=0.65\linewidth]{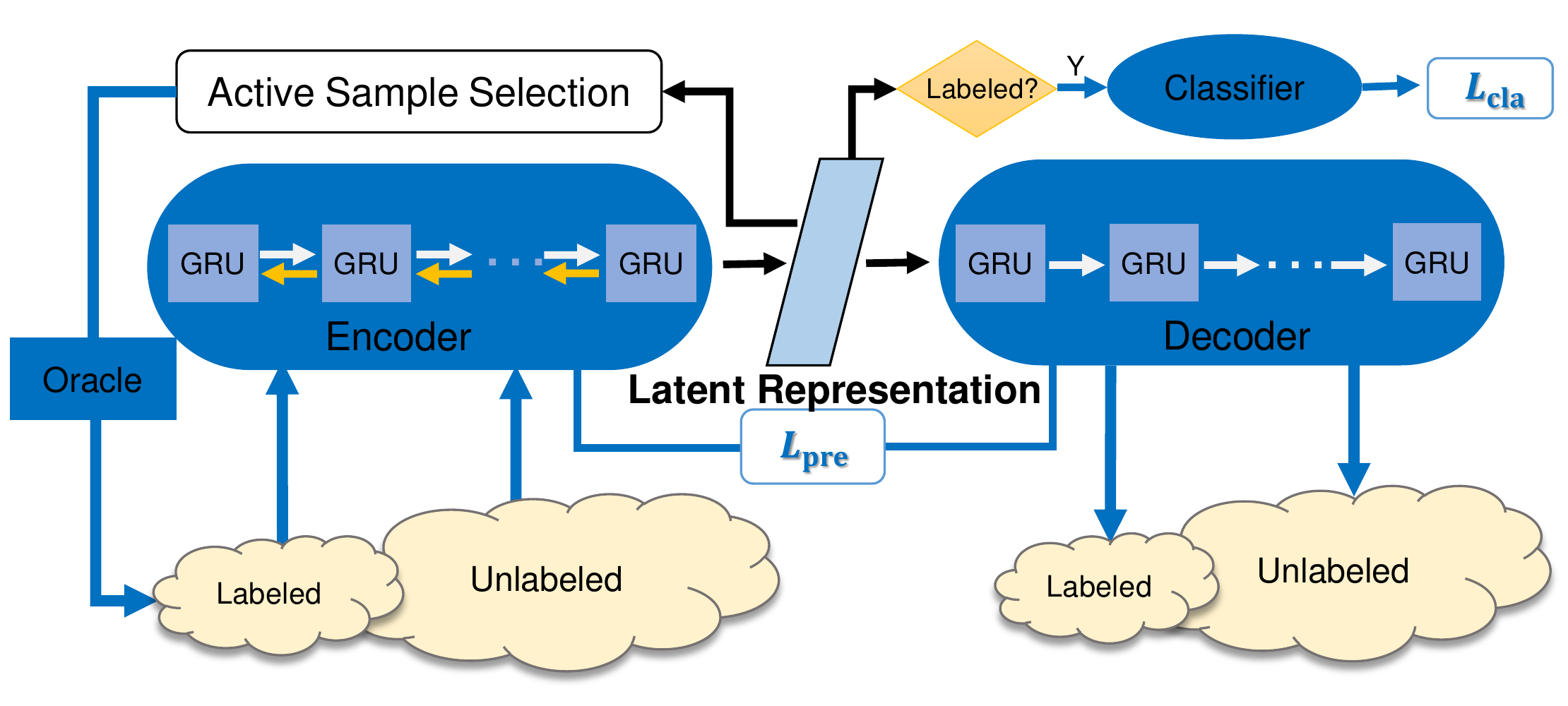} 
\end{center}
   \caption{SESAR system overview. A GRU based Encoder-Decoder structure is combined with a single layer Classifier through the latent representation between the Encoder and the Decoder. 
   The latent representation is used during training to reconstruct the sequences (unsupervised task) and to associate the sequences with their appropriate classes (based on labels). In addition, the proposed active learning strategy is using the latent representation for selection of sequences for annotations to enhance association with classes.
}
\label{fig:teaser}
\end{figure*}
Recognition of actions from spatial temporal sequences is a key component in ubiquitous applications such as action recognition of human movements, understanding subject interaction, scenario analysis and even correlating neuronal activity with behaviors. 
Significant progress have been made in human action recognition from video. Due to the variety of the informational cues in the video there is an assortment of methods focusing on different features to associate movements with particular labels of actions.
A direct approach is to attempt to perform action recognition from the image frames of the video (RGB) or even to consider image and depth information (RGB+D). While robust action recognition can be obtained from video, pixel data includes redundant misleading features. For example, features from the background of the video are often irrelevant to objects actions. Due to these factors, to achieve efficient performance, current state-of-the-art RGB based methods require an extensive supervised training with a comprehensive annotated dataset even for a handful of actions. 

to leverage the information and assign label for the action being performed by objects in the video frames sequence. 
Complementary to the direct consideration of video image frames as an input, markerless feature detector methods offer a rapid detection of cues in each frame. For common objects, such as people or their faces, these are standard and robust features (keypoints) that define pose estimation such as the skeleton joints or contours~\cite{rahmani2014hopc, wang2016temporal}.
The use of keypoints for action recognition is advantageous since they filter the unnecessary information and provide concise sequences from which actions can be extracted. 
Several systems have been introduced for action recognition from body skeleton keypoints, learning spatial and temporal relations for each sequence and associating action with it through fully supervised approach~\cite{du2015hierarchical, li2018co, shi2020skeleton}.
While the performance of supervised methods on datasets of human actions with cross view or cross subject variance have been shown to be effective such approaches require massive number of annotations in the training set, e.g., a typical benchmark for 60 cross subject actions (NTU RGB+D 60) includes 40,000 annotated sequences. The reliance on labels perplexes the possibility of scaling up the approaches to more actions and also makes inclusion of mislabeled sequences to be more probable and deteriorates performance. Indeed, it is often the case that the annotation is split between annotators and it is up to the annotator's interpretation to assign a meaningful label for a given sequence. 

Recent advances suggest a possibility to overcome the annotation requirement by the implementation of unsupervised action recognition system from keypoints~\cite{zheng2018unsupervised,su2019predict,suclustering}. These systems use an encoder-decoder to regenerate the sequences of keypoints. The encoder-decoder structure is found to self-organize its latent states to clusters which correspond to actions. While these methods appear to be effective and promising the association of clusters with actions still remains a step that requires labels. Indeed, to evaluate the performance of the unsupervised methods it is proposed to train a multilayer fully connected network or KNN classifier on the full training set~\cite{zheng2018unsupervised,su2019predict}.

To connect the unsupervised self-organization with the association of actions leads to the consideration of efficient strategies that will require only a few annotations. This directs the consideration of semi-supervised learning strategies and indeed recent approaches show that consideration of only a subset of annotated samples, e.g., from about 4,000 sequences to as low as 100 sequences is possible~\cite{si2020adversarial, lin2020ms2l}. The selection of the sequences to be annotated is performed such that  
a certain number of samples is evenly selected from each action or randomly selected without accounting for the actions classes. These selection strategies may not be optimal in the performance of action recognition in ubiquitous settings. For example, the even selection may not be possible in the case of recognition of novel actions since these are unavailable a prior. The random selection on the other hand can lead to uncertain performance. The generalization leads to the consideration of sampling strategies which are informed by the training process, i.e., active learning~\cite{settles2009active}.
Typical active learning approaches are `pool-based', where 
for a pool of possible candidates it is being evaluated which of them provides the most information for the training process. 

We therefore propose a novel pool-based active learning approach for Semi-Supervised Action Recognition (SESAR) that merges the self-organization of unsupervised methodology with the association of action classes. The pool with our approach is the full training set and we show that even a sparse selection of sequences for annotation (samples) from that pool, as low as $1\%$, the method is able to generate a robust representation of classes. 

Established active learning approaches typically consider different tasks than action recognition of sequences considered here. They typically focus on object recognition in images and as such, aim to select samples that most optimally represent the class that they belong to consider only uncertainty of samples. For recognition of action sequences, however, a more suitable selection would aim to find samples that will not only represent their own classes but will assist with representation of as many classes as possible, i.e., distinguish between classes. Therefore, SESAR includes this distinctive selection within its architecture by employing the uncertainty and similarity estimation. This goal is similar to adversarial active learning~\cite{sinha2019variational} which seeks to determine the distinctive samples, however, the advantage in SESAR is that it utilizes the latent space representation for the selection, while adversarial approaches would require training each time the selection is made.

In summary, we introduce a Sparse Semi-Supervised Action Recognition, SESAR, which implements an Active Learning approach for skeleton-based action recognition. Our main contributions are (i: propose a novel active learning approach for action recognition of spatiotemporal sequences which selects sequences for annotation according to the estimation of the effectiveness of the annotation of each sample with respect to its representation, similarity and uncertainty. 
(ii) The approach provides a general effective algorithm of combining unsupervised sequence reconstruction with classification through latent representation embedding to perform semi-supervised learning for sequences action recognition. (iii) We apply the approach to three common extensive benchmarks of skeleton-based human action recognition (UWA3D, NW-UCLA, NTU-RGB+D) and show that we achieve high efficiency performance for sparse selection of annotations, with improvement of $5-10\%$ than semi-supervised state-of-the-art methods and achieve better performance than established active learning approaches designed for other problems than action recognition of sequences. \\

\section{Related work}
\textbf{\textit{Supervised Action Recognition.}} A variety of \textit{supervised learning} skeleton-based action recognition approaches have been proposed. Earlier studies focused on computing local statistics from spatial salience and action motion associating them with actions~\cite{xia2012view, wang2013learning, vemulapalli2014human}. Recently, methods based on deep learning have been proposed. Such systems are structured with underlying Recurrent Neural Networks (RNN) or Convolutional Neural Networks (CNN) architectures. RNN based methods include an end-to-end hierarchical RNN~\cite{du2015hierarchical}, part-aware long short-term memory (P-LSTM) unit~\cite{shahroudy2016ntu}, or co-occurence LSTM~\cite{zhu2016co} using internal part based memory sub-cells with new gating strategies for the skeleton action recognition. CNN approaches propose to transform the skeleton sequences into color images which are then taken as an input into a network to extract features of spatial temporal information~\cite{liu2017enhanced, kim2017interpretable, ke2017new}. Furthermore, several recent approaches propose to add relations, e.g. bones, between skeleton joint information. Such an extension turns the underlying architecture into a Graph Convolutional Network (GCN) and enhances the performance of supervised action recognition~\cite{yan2018spatial, zhang2020context, cheng2020skeleton, li2020dynamic}. 

\textbf{\textit{Unsupervised Action Recognition}.}
While supervised methods have been shown to be effective, they are limited by a time consuming process of annotation, prone to ambiguity. These limitations lead to the consideration of \textit{unsupervised methods}. Previous work shows an encoder-decoder sequence to sequence (Seq2Seq) networks, initialized with random weights, can self-organize their latent state space to cluster sequences in a low-dimensional embedding space~\cite{su2019predict,farrell2019recurrent}. 
Such approaches do not require annotations. Instead, they seek to learn a representation to capture the long-term global motion dynamics in skeleton sequences through learning to reconstruct the same sequences as the input~\cite{zheng2018unsupervised}. Such a setting indicates the capability of the network to self-organize the sequences to cluster actions within its latent state space. Latest representation learning methods propose ways for extraction of self-organization functionality and the network architectures to perform high-precision recognition~\cite{su2019predict,suclustering}. In particular, GRU based encoder-decoder RNN pipeline was introduced, in which the decoder is weakened, such that the latent representation of the encoder, in particular the last state of it, will be promoted to represent different actions as different classes. The system was evaluated using a KNN classifier and showed promising clustering and classification performance. The KNN classifier is comparable to supervised methods and needs to be trained on the full dataset. Our goal is therefore to consider approaches that rather than performing association as a post-processing step will incorporate this step within the network itself. Furthermore, it is key to require a minimal number of annotations (sparse semi-supervision) chosen in an active manner (active learning) utilizing the self-organization functionality to remain as close as possible to the unsupervised setting.

\textbf{\textit{Semi-supervised Action Recognition}.}
A few recent approaches have been proposed to perform \textit{semi-supervised} action recognition, i.e., to add a classifier to the encoder-decoder architecture and to perform classification task is our setup. 
In~\cite{si2020adversarial} it was proposed to utilize the neighbors assumption along with an adversarial approach (ASSL) to distinguish feature representation for labeled and unlabeled sequences. This strategy turns out to be  advantageous for effective performance and requires a smaller set of annotations ($5\%-40\%$) than the full training set. One constraint of the approach is that it selects the sequences for annotation evenly from each class assuming that the classes are already known. To amend this limitation, augmentation is utilized in~\cite{lin2020ms2l} ($MS^2L$) to learn a more robust representation such that unsupervised action recognition is performed on an augmented datatset with the representation and then sequences are selected for annotation randomly assuming a favorable distribution of the latent states space. The approach does not require any prior knowledge of the classes and achieves effective performance on human movement action recognition benchmarks. Since the approach requires augmentation, it appears that to achieve optimal performance it requires a substantial amount of annotated sequences exceeding the requirement of previous approaches.

\textbf{\textit{Active Learning.}}
To select sequences for annotation and to perform class association with only a few samples, there is a need for a novel approach that combines the unsupervised learning with an active selection of samples for annotation. Indeed, fundamental methodology for \textit{Active Learning (AL)} has been broadly established. Active Learning methods are typically belong to three categories: (i) sample synthesis, (ii) query by committee, and (iii) pool based sampling with statistical methods \cite{settles2009active, cohn1996active, hanneke2014theory}. The methods in the pool based category are more common in classical machine learning techniques, such as SVM, since they are not required to synthesize samples and can sample data points from the entire dataset. The selection is guided according to metrics which describe the uncertainty or the representation of particular features and the selection procedure is based on a statistical sampling, such as, information theoretical methods, ensemble approaches, and uncertainty based methods~\cite{xu2003representative, nguyen2004active, schohn2000less, tong2001support}. For deep-learning approaches, active learning is being adapted to support sample selection according to transformed features in the latent space representation. These approaches learn a metric to measure the uncertainty of a sample~\cite{gal2017deep} through Bayesian CNN, or find representatives of clusters~\cite{coletta2019combining}. The intent of such approaches is to cover as many features as possible with each selected sample. Indeed, such AL approaches were applied in batch setting to enhance the performance of CNN trained on image classification~\cite{sener2017active} or to leverage the information from known classes to enhance the performance of new classes~\cite{coletta2019combining}. 

Inherent deep adversarial AL methods have been proposed as well. These methods propose to learn a latent representation from the full training dataset first and then to train a discriminator for the decision of whether to annotate a particular sample. Such methods were successfully applied to the `slot filling' and 'captionining' tasks~\cite{deng2018adversarial} and for image classification task~\cite{sinha2019variational}. 

While these works consider similarity as the main measure, for skeleton-based action recognition it is required to consider as many metrics as possible. It is also necessary to perform batch sampling of sequences and therefore to take into account reduction of the dimension of the data. In SESAR, we address these unique aspects of action recognition, as described in the Methods section below, and compare the performance on benchmarks of skeleton-based action recognition. Our experiments show that SESAR can achieve an enhanced action recognition performance and would require less samples when compared with AL methods not specifically designed for action recognition. 

\section{Methods}
\begin{figure}[t]
\begin{center}
   \includegraphics[width=0.95\linewidth]{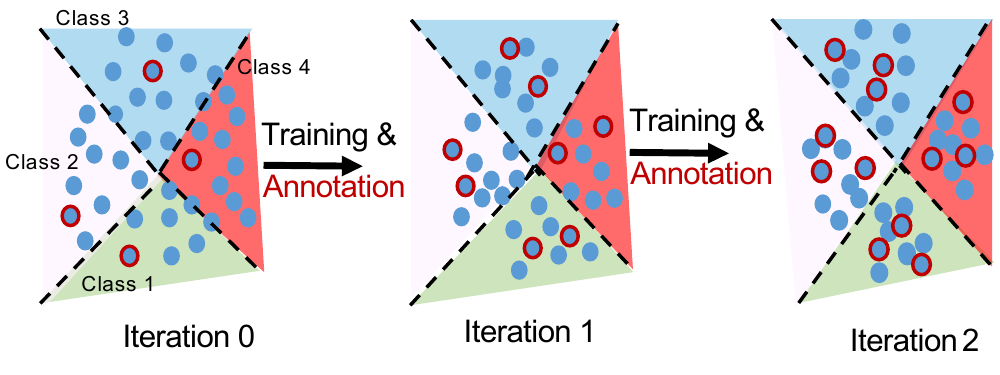}
\end{center}
   \caption{Illustration of Active Learning. Three iterations of the active selection of samples for annotation are shown. In each iteration, clusters are formed in the latent representation space. Samples for annotation are chosen from each cluster according to different active learning strategies. The process is repeated for multiple iterations according to the overall allowed extent of annotation.}
\label{fig:labelselection}
\end{figure}
We describe the two main components of the proposed SESAR system in this section: (i) Semi-supervised action recognition system that combines a classification task with an unsupervised regeneration task for action reconstruction. (ii) Implementation of Active Learning which selects the sequences to be annotated. An overview of SESAR system architecture is depicted in Figure~\ref{fig:teaser}. 

\textbf{\textit{Data Preparation}} SESAR works with datasets of multi-dimensional time series specifying the coordinates of body keypoints at each given time. We denote the times series as $\mathcal{X} = \{X_{u}\bigcup X_{l}\}$, with $X_u$ representing the keypoints in the unlabeled set, and $X_l$ in the labeled set. A sample $\mathbf{x} \in \mathcal{X}$ is represented as a sequence $\mathbf{x} = [x_1, x_2, ..,x_t, ... x_T]$, where $x_t$ is the coordinates of the keypoints  at time $i$,  $x_t \in R^{N \times D}$, here $N$ is the number of keypoints, $D$ (typically $D=3$) is the dimension of the keypoints. The number of keypoints, $N$,  is expected to vary  for different dataset. For the datasets with keypoints obtained from video frames recorded from multiple views (e.g. NW-UCLA, UWA3D) we follow the procedure of transforming them to a view invariant representation~\cite{su2019predict, shotton2011real}. 

\textbf{\textit{Model Structure and Objective}} SESAR model includes three major components: The encoder, the decoder and the classifier, as depicted in Fig.~\ref{fig:teaser}. The encoder includes Gated Recurrent Units (GRU) and receives  $\mathbf{x} \in \mathcal{X}$ as its input. The latent state within the encoder accumulates, at time step $t$, the characteristic sequence dynamic properties from time $0$ to time $t-1$, such that at time $T$, the last latent state of encoder contains the dynamic properties of the whole sequence. We use bidirectional GRUs, which is broadly used in action recognition tasks to account for both forward and backward dynamics in the sequences~\cite{su2019predict, si2020adversarial, zheng2018unsupervised}. The last latent state of the encoder denotes as $H_i$ for sample $i$, is considered as a coordinating representation for the three components.

In particular, it coordinates two complementary paths. The first path is the flow from the encoder to the decoder, where the latent representation $H_i$ is transferred to the decoder. In this flow the input sequence is being reconstructed by the decoder using the reconstruction loss, i.e., minimization of the distance between the generated $\mathbf{\hat{x}_i}$ sequence by the decoder and the input sequence $\mathbf{x}$. Here we use $L_1$ loss as reconstruction loss
\begin{align}\label{loss:re}
\mathcal{L}_{re} = |\mathbf{x}_i - \mathbf{\hat{x}_i}|.
\end{align}

The second path is the flow from the encoder to the classifier. The classifier that we use is a single fully connected layer. The benefit for such a simple classifier is that in the sparse annotation scenario we aim to avoid overfitting of the classifier and position most of the nuanced classification ability in the latent representation. Indeed, in such a way, the latent representation is organized according to both features of reconstruction and classification. Such penalization strategies were proposed in prior work, where to enhance the encoder amenability for classification it was found necessary to `weaken'  the decoder~\cite{su2019predict}, or inpaint the decoder input ~\cite{zheng2018unsupervised} and we penalize the decoder as described in~\cite{su2019predict}. 
The loss function for the classifier is defined as
\begin{equation}\label{loss:cla}
\mathcal{L}_{cla} = \sum_{j=1}^{C}-y^j\log (p_j(\mathbf{x}_i)).
\end{equation}
where $p_j(\mathbf{x_i})$ is the probability of an output sample $i$ to belong to class $j$ and $C$ is the number of classes.

The combination of above two losses translates to the following training procedure for each sample $\mathbf{x}_i$
\begin{equation} \label{loss:i}
    \mathcal{L}_{\mathbf{x_i}}=\left\{ \begin{array}{ccl}
     \mathcal{L}_{re} + \mathcal{L}_{cla} & \mbox{if} & \mathbf{x_i} \in X_l, \\
      \mathcal{L}_{re} & \mbox{if} &\mathbf{x_i} \in X_l,
\end{array} \right.
\end{equation}
and the full model is trained according to the total loss of
\begin{equation}\label{eq:totalloss}
    \mathcal{L} = \sum_{\mathbf{x_i} \in \mathcal{X}}~\mathcal{L_{\mathbf{x_i}}}.
\end{equation}

\textbf{\textit{Leveraging of Clustering for AL}}
The setup of our proposed AL is to first learn a latent representation for sequence samples using an unsupervised approach and then select according to this learned latent representation samples which will be passed to an `Oracle' for annotation. The annotated samples are added to the labeled set and the model is re-trained with all labeled samples using a classifier. This process enhances the overall classification performance and also the latent representation by bringing samples that belong to the same class closer and distancing the distances in the latent representation space between the classes. The general procedure is illustrated in Fig.~\ref{fig:labelselection}, where blue circles represent samples located in the latent space, blue circles with red  borders are samples selected for annotation. At Iteration $0$ the latent representation embedding is achieved with the unsupervised reconstruction only, clusters are computed (dashed lines) and samples according to a particular metric (e.g., near the center of each cluster) are selected for annotation. After Iteration $0$, the model is trained with the full loss defined in Eq.~\ref{eq:totalloss}. and with the newly learned latent representation, which becomes more enhanced and contains clearer clusters, another set of samples is chosen for annotation. The process is then repeated in an iterative fashion.

\textbf{\textit{Sample Representation Measure}}
To form the clusters we propose to apply a clustering method to the encoder latent representation ($H_i$). We use the \textit{KMeans} clustering to form samples into a series of clusters $\mathbf{\Tilde{C}_i}$. Other approaches, such as Hierarchical clustering~\cite{johnson1967hierarchical}, could also be applied in a similar fashion. To compute the distance of each sample with each cluster in the latent space, we compute $d_i^{\Tilde{c}}$ which is the distance of the sample $i$ to a designated cluster $\Tilde{c}$ (label as $l_i^{\Tilde{c}}$). The number of samples to annotate in a cluster is $n^{\Tilde{c}} = \textit{Sample percentage} \times N^{\Tilde{c}}$, where $N^{\Tilde{c}}$ is the total number of samples in the cluster $\Tilde{c}$. The distance allows us to select samples according to their position in the cluster. Typically, samples that are in the center of the cluster would better represent the cluster than those that are distant from the center. We thereby propose an AL strategy, named KMeans-Top (\textbf{SESAR-KT}), which selects samples for annotation that are the closest to each cluster center. In practice, we set the number of clusters formed in each iteration to be a maximum of $2C$ for stability and variability, where $C$ is the number of classes in the considered dataset.

\textbf{\textit{Sample Similarity and Uncertainty Measures}}
One potential limitation of the KT selection strategy is that when the selection is performed with multiple iterations, as illustrated in Fig.~\ref{fig:labelselection}, samples that are close to the cluster center in the initial iterations are likely to remain close to the cluster center at the subsequent iterations. Hence, the selection procedure is likely to select similar samples across iterations of KT. We thereby propose to incorporate a selection strategy that is balanced, such that the selected sample for annotation is both in proximity to the center of the cluster and is distant (dissimilar) enough from samples that are annotated, i.e., $\mathbf{x_u}$ and $\mathbf{x_l}$.  We propose to measure the similarity with the Jensen–Shannon divergence~\cite{manning1999foundations} of class probability distribution between the labeled samples $\mathbf{x}_l$ and the unlabeled samples $\mathbf{x}_u$ defined as follows
\begin{equation}
S(\mathbf{x_l}, \mathbf{x_u}) = \frac{1}{2}D(p(\mathbf{x}_l)|| p_z) + \frac{1}{2}D(p(\mathbf{x}_u)||p_z),\\
\end{equation}
where $D$ is the KL divergence between two distributions, $p$ is softmax output from the classifier.
\begin{equation}
    p_z = \frac{1}{2}(p(\mathbf{x}_l) + p(\mathbf{x}_u))
\end{equation}
\begin{equation}
    D(p_1||p_2) = \sum p_1\log(\frac{p_1}{p_2}) \
\end{equation}
We compute the similarity between the unlabeled and the labeled samples in each cluster. When there are no labeled samples in a cluster, the sample with the smallest distance to the cluster center is chosen. 

In addition, we consider the uncertainty measure of samples. 
We follow the uncertainty computing procedure of~\cite{activ2012, gal2017deep}, in which Variance Ratio (VR) and Entropy (EP) metrics are used to estimate uncertainty. Selection of samples according to the uncertainty is a common practice in AL. Selection is typically performed sequentially with a single sample selected each time. In our proposed approach multiple samples are selected at once for adequate representation of clusters. We implement a softmax output from the classifier output layer which provides samples class probability distribution, such that a sample $i$ will have $\mathbf{p_i} = [p_i^1, p_i^j, ... p_i^C]$ probability to belong to each class. We define the uncertainty vector for whole population as $\mathbf{P} = [p_1^{c*}, p_2^{c^*} ..., p_i^{c^*}]$ for all samples in training set, where the class which corresponds to highest probability is denoted as $c^*$. According to the VR measure, sampling $M$ samples results in selecting $M$ samples corresponding to the smallest M value in $\mathbf{P}$. The entropy-based approach computing the entropy of prediction for each sample $i$, as 
$
    H(X_i) = -\sum_{c=1}^{C}(p_i^j\log (p_i^j)).
$
EP strategy selects $M$ samples with the maximal $H$ (entropy) values.
With the combined similarity and uncertainty selection we take into account both the association of samples to clusters, the information provided by the unsupervised learning, and also the uncertainty of each sample to complement that information in the annotation. We call this selection process as \textbf{SESAR-KJS}. Further details regarding the procedure are shown in Algorithm~\ref{uclid}.

\begin{algorithm}
\caption{Active Selection of Samples Leveraging Unsupervised Clustering}\label{uclid}
\begin{algorithmic}[1]
\State Input: unlabeled and labeled Samples $\mathbf{X_u}$, $\mathbf{X_l}$, $p$ sample probability, $M$ number of clusters.
\State $\mathbf{H_u}, \mathbf{H_l} = \mathit{Encoder}(\mathbf{X_u}), \mathit{Encoder}(\mathbf{X_l})$
\State $\mathbf{\Tilde{C}}$, $d_{\mathbf{x_i}}^{\Tilde{c}}$, $n^{\Tilde{c}}$ = $\mathit{KMeans}(\mathbf{H_u}, \mathbf{H_l} , M)$
\For{Cluster $\mathbf{\Tilde{C_i}}, i = 1:M$}
\If {$\mathbf{X_l} \cap \mathbf{\Tilde{C_i}} = \emptyset$}
   \State $s = \mathit{argsort}_{x_i \in \mathbf{\Tilde{C_i}}}(d_i^{\Tilde{c}})$ (ascending sort)
\Else
   \State $sim(\mathbf{x_i}) = min_j(S(x_i, x_j))$, $x_i \in \mathbf{X_u} \cap \mathbf{\Tilde{C_i}} $, $ x_j \in \mathbf{X_l} \cap \mathbf{\Tilde{C_i}}$
   
   \State $\mathbf{S_0} =\{sim(\mathit{x_i})|~x_i \in \mathbf{X_u} \cap \mathbf{\Tilde{C_i}}\}$
   \State $x_0 = \mathit{argsort}_{x_i \in \mathbf{X_u} \cap \mathbf{\Tilde{C_i}}}(\mathbf{S_0})$ (descending sort)
   \State $\mathbf{X_0} = x_0[:2*n^{\Tilde{c_i}}]$
   \State $u_{\mathbf{x_i}} = Uncertainty(x_i), ~x_i \in \mathbf{X_0}$
   \State $s = \mathit{argsort}_{\mathbf{x_i} \in X_0}(u_{\mathbf{x_i}})$ (descending sort)
\EndIf
\State $\mathbf{X_l} = \mathbf{X_l} \cup s[:n^{\Tilde{c_i}}]$
\EndFor
\label{al:algorithm}
\end{algorithmic}
\end{algorithm}

\textbf{\textit{Action Recognition with Common Active Learning Methods}}
To contrast the AL methods that leverage the clustering property of the latent representation achieved by unsupervised learning (SESAR-KT, SESAR-KJS) we also examine the application of current state-of-the-art AL methods directly on the latent representation. In particular, we implement and examine pool methods of Uniform sampling (SESAR-U), Core-Set (SESAR-CS)~\cite{sener2017active}, and Discriminator based selection (SESAR-DIS)~\cite{deng2018adversarial, sinha2019variational}. SESAR-U performs a random selection from the unlabeled pool. SESAR-CS aims to minimize the cover range of each selected sample in order to make the selected samples cover the whole space when a batch selection is performed. SESAR-DIS is based on an additional discriminator to distinguish if a sample is labeled or not. Samples with a high probability of being unlabeled are selected for annotation.

\begin{figure*}
\begin{center}
  \includegraphics[width=0.8\linewidth]{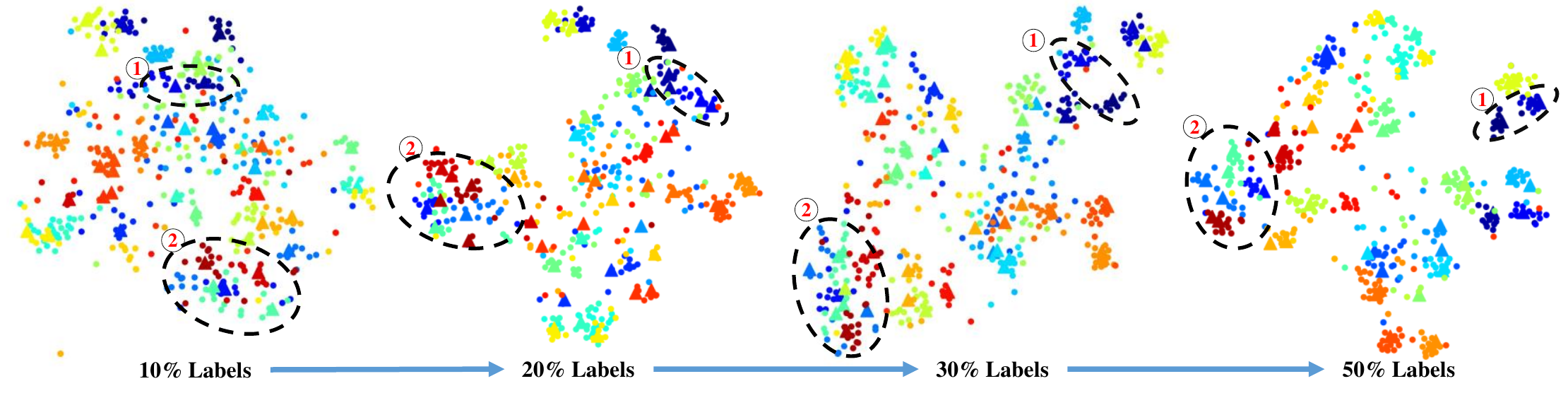}

   \caption{t-SNE embeddings of the latent representation across three training iterations. Colors correspond to action classes. Iteration $0$ is initialized with $10\%$ annotated samples, where in each iteration additional $10\%$ of samples are annotated. Two sample sets are denoted by dashed circles to show how the samples form distinct clusters that correspond to distinct actions.}
   
\end{center}
\label{fig:ember}
\end{figure*}
\section{Results}
\textbf{\textit{Datasets.}}
We evaluate the performance of SESAR on three common benchmark datasets, \textit{Multiview Activity II (UWA3D)}~\cite{rahmani2014hopc}, \textit{North-Western UCLA (NW-UCLA)}~\cite{wang2014cross}, \textit{NTU RGB+D 60}~\cite{shahroudy2016ntu}. These three datatsets constitute an extensive validation since containing the different number of actions, and cross-view and cross-subject sequences. In particular,
\textbf{\textit{UWA3D}} contains 30 human action categories. Each action is performed 4 times by 10 subjects recorded from frontal, left, right, and top views. We selected the first two views as training sets and the third view as the test set, which appears to be a more challenging task according to the performance scores of related work~\cite{su2019predict, zhang2017view}).
\textbf{\textit{NW-UCLA}} dataset is captured by three Kinect Automated Home-Cage Behavioral cameras containing depth and human skeleton data from three different views. The dataset includes 10 different action categories performed by 10 different subjects repeated from 1 to 10 times. We use the first two views as the training set, the last view (V3) as the test set following the same procedure as in~\cite{su2019predict, liu2017enhanced, wang2014cross}.
\textbf{\textit{NTU RGB+D 60}} includes both video sequences and skeleton sequences performed by 40 different subjects recorded using 3 different cameras across different views. The dataset including 60 different classes. We evaluate the performance in both the cross-view setting (see Supplementary Materials) and the cross-subject setting. For the cross-view setting, samples from cameras 2 and 3 are used for training, and samples from camera 1 are used for testing. Cross-subject setting splits subjects into 2 groups, 20 subjects are used for training and the other 20 subjects are used for testing, which is a harder task compared to cross-view, especially for unsupervised methods or semi-supervised~\cite{su2019predict, si2020adversarial}.
\begin{figure}[t]
\begin{center}
   \includegraphics[width=1\linewidth]{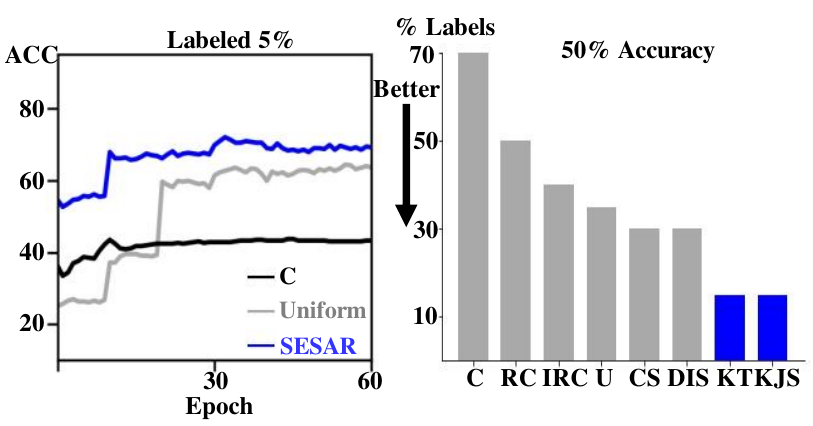}
\end{center}
   \caption{Left: Training trajectory with $5\%$ labeled samples using C (black), U (short for SESAR-U in gray), and KT (short for SESAR-KT in blue) methods on UCLA dataset. Right: Label requirement for achieving $50\%$ percent accuracy on \textit{UWA-3D} dataset. Comparisons are made among all C, RC, IRC, U (short for SESAR-Uniform), DIS (short for SESAR-DIS), KT (short for SESAR-KT), and KJS (short for SESAR-KJS)}.
\label{fig:trajectory}
\end{figure}
\textbf{\textit{Implementation Details}}
For our experiments, the encoder constitutes three-layers of bi-GRU cells, with 1024 hidden units for each direction. Hidden units from both directions are concatenated to a 2048 dimensional latent representation, then taken as an input to the decoder. The decoder is a uni-GRU with a hidden size of 2048. The classifier, a single fully connected layer receives the latent representation as an input and computes class probabilities for each sample. Adam optimizer is used for optimization. The learning rate is set to be 0.0001, with a 0.95 decay rate every 1000 iterations for UWA-3D and NW-UCLA dataset, 500 iterations for NTU RGB+D 60 dataset.

To test and validate the effectiveness of SESAR, we compare its performance with various semi-supervised methods (SSL) in Tables~\ref{resultuwa3d}, \ref{resultucla}, \ref{resultntu}. In particular, we compare the performance of baseline methods (C, RC and RIC) which are ablated versions of the SESAR. \textbf{C} corresponds to the classification only approach, with a three-layer bidirectional GRU encoder and a single fully connected classification layer. \textbf{RC} corresponds to the reconstruction and classification model implemented at the same time, corresponding to associating an unsupervised method to classes with a limited number of samples. Another variation of such network is \textbf{RIC} in which the unsupervised model is first trained on the reconstruction task, after which a classifier is added and the simultaneous learning of reconstruction and classification is performed. In addition, we compare the performance with recently introduced methods for semi-supervised learning for skeleton-based action recognition (\textbf{ASSL}~\cite{si2020adversarial} and \textbf{MS$^2$L}~\cite{lin2020ms2l}). 

Along with the specification of the percentage of the labels selected for annotation we also specify the actual number of samples. We focus on evaluating the performance in a sparse sampling scenario where in UWA3D $5\%-50\%$ ($25-250$) samples are annotated, in NW-UCLA $5\%-40\%$ ($50-400$) are annotated and in NTU-RGB+D 60 $1\%-10\%$ ($400-4K$) samples are annotated. The results for the baseline methods and the SESAR methods are reported as mean results to exclude variation in sample selection, which is unavoidable, especially in sparse selection.

\textbf{\textit{Importance of the Latent Space}}. We observe that classification only (C) performs with the lowest success in most of the tests. When the classification is augmented with reconstruction (RC) we observe a boost in the performance. Initialization with an initially trained reconstruction model (RIC) improves the action recognition even further. These results indicate that leveraging the unsupervised reconstruction and its clustering property is an advantageous approach. This is supported by inspection of the t-SNE embedding of the latent representation depicted in Fig.~\ref{fig:ember}, which shows that training leveraging clustering (SESAR-KT) succeeds to form clusters of samples that are clearly identified with a labeled action.

To elucidate how many annotated samples each method would require, in Figure~\ref{fig:labelselection} (right) we test the required percentage of labeled samples for \textit{UWA-3D} dataset to achieve $50\%$ accuracy. While the sole classification approach (C) needs for $70\%$ of the samples to be annotated, SESAR-KT and SESAR-KJS require only $15\%$. When SESAR AL approaches that do not leverage clustering are considered they appear to require $30\%$.

\textbf{\textit{Performance of SESAR AL+K Methods}}
Indeed, across datasets, we observe that SESAR methods leveraging clustering, i.e., SESAR-KT and SESAR-KJS (AL+K), consistently perform more optimally than other approaches. In the scenario of sparse samples selected for annotation (the first column of Tables~\ref{resultuwa3d}, \ref{resultucla}, \ref{resultntu}), AL+K methods outperform SSL approaches by $5\%$ or more. Notably, for an extremely low number of samples, i.e., the first column of UWA3D, in which only 25 samples are selected for annotation, SESAR-KJS achieves the best performance, while on other datasets SESAR-KT achieves the best performance. This appears to indicate the particular use cases of the two approaches.  SESAR-KJS appears to be suitable for extremely sparse annotation while SESAR-KT is suitable for moderately sparse annotation. It is expected by the definition of these two methods, as SESAR-KJS incorporates both similarity and uncertainty and thus expected to perform well with only a few samples, while SESAR-KT may require more samples.

When the number of annotated samples increases, i.e. the last columns of Tables~\ref{resultuwa3d}, \ref{resultucla}, \ref{resultntu}, we observe that the performance of all considered SESAR methods are improving. Solely active learning based methods, AL, such as SESAR-DIS and SESAR-CS, achieve similar performance or perform better than the AL+K methods. These results demonstrate that the precise selection of samples is highly important in the case of sparse sampling and is relaxed when it is allowed to select more sequences for annotation. We further look into the training process to show the effectiveness of SESAR on the initial sparse annotated set. In Fig.~\ref{fig:trajectory} (left), we show that with $5\%$ SESAR-KT is quickly gaining accuracy within the first few epochs while other approaches such as a sole classifier remain constant during training while SESAR-Uniform requires many epochs to gain accuracy. 

\textbf{\textit{Comparison with SOTA Semi-Supervised Methods}} 
 Comparison with current semi-supervised methods for skeleton-based action recognition(ASSL~\cite{si2020adversarial} and MS$^2$L~\cite{lin2020ms2l}) shows that in most of the scenarios SESAR outperforms these methods. As in the comparison between SESAR AL and SESAR AL+K methods we observe that as the number of samples increases the gap between SESAR and ASSL and MS$^2$L is diminishing. In particular, for the NTU dataset with $10\%$ of annotated samples (4,000 samples), we obtain that these methods perform slightly better than SESAR. This is expected since the expanded influence of samples with annotation by considering the neighborhood of those samples or augmenting samples themselves will be further enlarged with so many annotations.

\begin{table}
\vspace{0pt}
\centering
\begin{tabular}{@{}clccccrrcrrr@{}}\toprule
& \multicolumn{4}{c}{\textbf{UWA3D VIEW3}} \\
\midrule
&\% Labels & $5\%$ & $10\%$ & $20\%$ &$50\%$ \\
&\#Labels  & $25$ & $50$ & $100$ &$250$\\ \midrule
 \multirow{3}{*}{SSL}
 &C &18.3&21.9&32.1&44.3\\
 &RC &19.5&30.0&26.9&46.3\\
 &IRC &20.0&36.4&37.6&51.1\\
\midrule
\multirow{3}{*}{AL(our)} 
&SESAR-DIS&18.5&29.2&40.9&55.6\\
&SESAR-U &21.8&31.3&41.0&55.8\\
&SESAR-CS&26.9&\textbf{37.1}&41.2&55.8 \\

\midrule
\multirow{2}{*}{AL+K(our)}
&{SESAR-KT} &{22.8}&{34.6}&{\textbf{51.8}}&{58.8}\\
&{SESAR-KJS} &{\textbf{28.3}}&{36.0}&{49.5}&{\textbf{59.5}}\\
\bottomrule
\end{tabular}
\caption{Performance of different semi-supervised approaches (top), SESAR with STOA AL methods (middle) SESAR with AL+K methods (bottom) on UWA3D dataset.}
\label{resultuwa3d}
\end{table}

\begin{table}
\vspace{0pt}
\centering
\begin{tabular}{@{}clccccrrcrrr@{}}\toprule
& \multicolumn{4}{c}{\textbf{NW-UCLA}} \\
\midrule
& \% Labels & $5\% $ & $15\%$ & $30\%$ &$40\%$\\
& \# Labels & $50$ & $150$ & $300$ &$400$\\ \midrule
 \multirow{5}{*}{SSL}
 &C &44.6&56.0&70.9&72.9\\
 &RC &51.5&63.1&77.0 &76.7\\
&ASSL\cite{si2020adversarial} &52.6&74.8&78.0&78.4\\
&RIC & 56.2 &70.5&73.5&81.0\\
&MS$^2$L\cite{lin2020ms2l} &\multicolumn{2}{c}{~--~~60.5~~--}&--&--\\
\midrule
\multirow{3}{*}{AL(our)} 
&SESAR-DIS &55.3&73.3&\textbf{80.3}&83.3\\
&SESAR-U &62.7&74.0&77.9&80.3\\
&SESAR-CS&\textbf{63.9}&71.5&77.5&82.3\\
\midrule
\multirow{2}{*}{AL+K(our)}
& {SESAR-KJS} & {58.1}&\ {76.6}& {\textbf{80.0}}&\ {\textbf{85.0}}\\
& {SESAR-KT} & {\textbf{63.6}}& {\textbf{76.8}} &\ {77.2} &\ {78.9}\\
\bottomrule
\end{tabular}
\caption{Performance of different semi-supervised approaches (top), SESAR with STOA AL methods (middle) SESAR with AL+K methods (bottom) on NW-UCLA dataset.}
\label{resultucla}
\end{table}

\begin{table}
\vspace{0pt}
\centering
\begin{tabular}{@{}clccccrrcrrr@{}}\toprule
& \multicolumn{4}{c}{\textbf{NTU RGB+D 60 Cross Subject}} \\
\midrule
&\% Labels & $1\%$ & $2\%$ & $5\%$ &$10\%$ \\
& \# Labels & $400$ & $800$ & $2K$ &$4K$\\ \midrule
 \multirow{5}{*}{SSL}
 &C &21.8&37.2&49.6&56.7\\
&MS$^2$L\cite{lin2020ms2l} &33.1&--&--&\textbf{65.2}\\
&RC &33.8&41.6&47.8&60.0\\
 &IRC &36.7&42.7&53.9&61.2\\
&ASSL\cite{si2020adversarial} &--&--&57.3&64.3\\
\midrule
\multirow{3}{*}{AL(our)} 
&SESAR-CS &17.6&23.1&37.0&49.6\\
&SESAR-DIS &34.9&39.5&53.8&60.4\\
&SESAR-U &36.1&42.5&53.9&60.8\\
\midrule
\multirow{2}{*}{AL+K(our)}
&SESAR-KJS & 38.2& 45.0&{\textbf{57.8}}&{62.9}\\
&{SESAR-KT} &{\textbf{41.8}}&{\textbf{46.1}} & {55.0} &{58.2}\\
\bottomrule
\end{tabular}
\caption{Performance of different semi-supervised approaches (top), SESAR with STOA AL methods (middle) SESAR with AL+K methods (bottom) on NTU RGB+D 60 dataset.}
\label{resultntu}
\end{table}

\section{Conclusion}
We introduce a novel Semi-Supervised Active Learning approach (SESAR) for recognition of action in sequences. The approach connects unsupervised learning with a sparse active selection of sequences for annotation and boosts the action recognition performance. We apply our approach to skeleton-based human action recognition benchmarks and compare the performance with current semi-supervised methods that do not employ unsupervised learning. We show that our proposed methods outperform these approaches in sparse annotation scenarios, \eg, when only a handful of samples is selected for annotation.

\end{document}